%% file: main.tex
\documentclass[runningheads]{llncs}
\usepackage{graphicx} 
\usepackage{todonotes}
\usepackage{parskip}
\usepackage{booktabs}
\usepackage[export]{adjustbox}
\usepackage{tabularx}
\usepackage{caption}
\usepackage{siunitx}
\usepackage{lipsum}
\usepackage{url}
\usepackage{hyperref}
\usepackage{float}
\usepackage{multirow}
\setuptodonotes{inline}
\usepackage{cleveref}
\title{LETT-NeXt: A Lightweight RECIST-Guided Model for 3D CT Lesion Segmentation}
\titlerunning{LETT-NeXt for CT Lesion Segmentation}
\author{
Sebastian~Aas\orcidID{0009-0006-2453-1289} \and \\
Elias~Stenhede\orcidID{0009-0005-2654-4553} \and \\
Arian~Ranjbar\orcidID{0000-0002-0422-2255}
} 

\date{\today}

\institute{
Medical Technology \& E-health, Akershus University Hospital, Lørenskog, Norway\\
Faculty of Medicine, University of Oslo, Oslo, Norway\\
\texttt{sebastian.aas@ahus.no}
}

\begin{document}

\maketitle
\input{sections/abstract}
\begin{figure}[h]
    \centering
    \includegraphics[width=\linewidth]{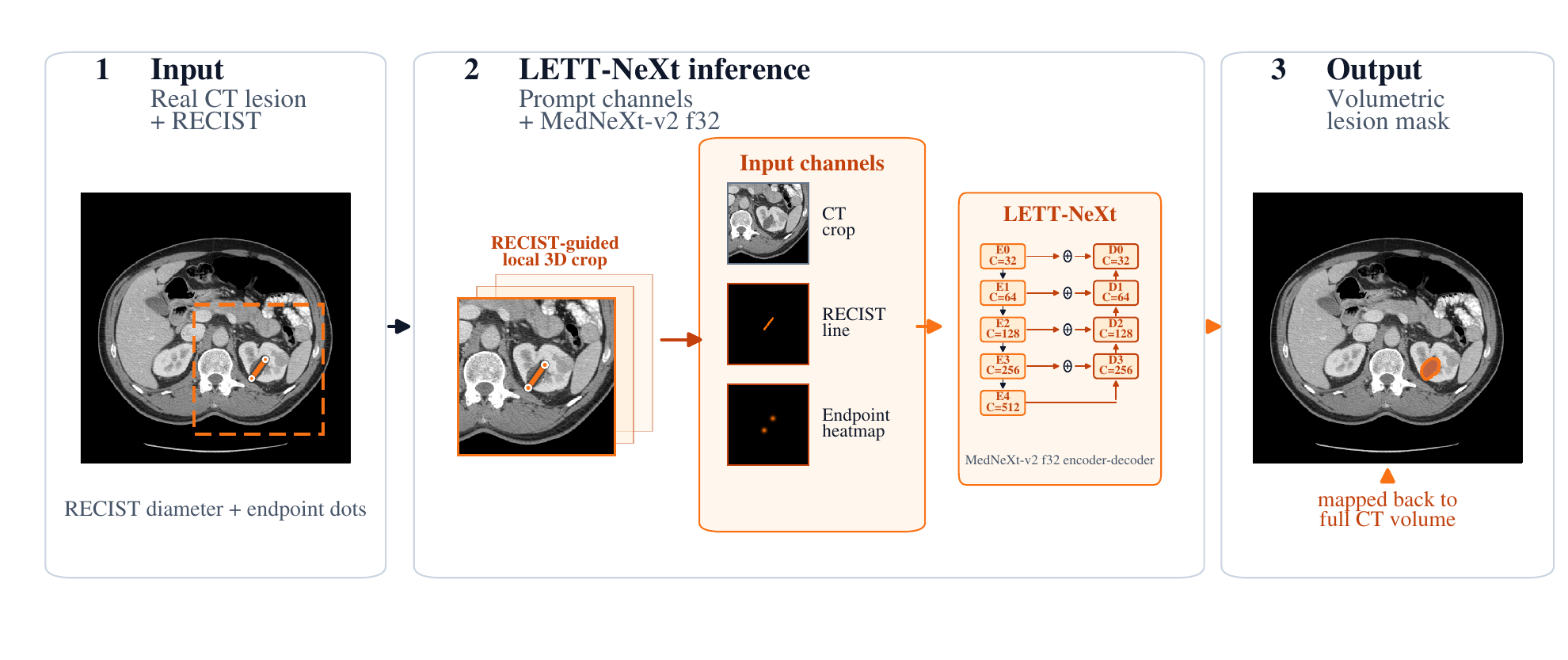}
    \caption{
        LETT-NeXt converts a RECIST marker into a volumetric lesion mask.
    }
    \label{fig:visual-abstract}
\end{figure}
\input{sections/introduction}
\input{sections/methods}
\input{sections/experiments}
\input{sections/results}

\input{sections/ablations}
\input{sections/discussion}

\input{sections/appendix}

\bibliographystyle{plain}
\bibliography{references}
\end{document}

%% file: sections/abstract.tex
\begin{abstract}
RECIST diameter measurements are widely used for tumor response assessment, but they provide only a limited 2D description of lesion extent. We present LETT-NeXt, a lightweight RECIST-guided model that predicts 3D lesion masks from CT volumes and RECIST markers for the CVPR 2026 Foundation Models for Pan-cancer Segmentation in CT Images competition. LETT-NeXt extracts a RECIST-centered regional crop, encodes the RECIST line and endpoints as two prompt channels, and concatenates them with the CT input. A compact MedNeXt-v2 encoder--decoder predicts the lesion mask, followed by prompt-aware component selection and adaptive AutoZoom inference. On the public validation set, LETT-NeXt achieved a Dice Similarity Coefficient (DSC) of 79.4 $\pm$ 10.1 and a Normalized Surface Dice (NSD) of 72.3 $\pm$ 16.2. On the hidden test set, it achieved a DSC of 73.9 and an NSD of 67.3, corresponding to a challenge score of 70.6\%. On the public validation mirror, LETT-NeXt completed CPU inference in 6.9 $\pm$ 3.0 s per case with a peak memory use of 3.6 GB. Code is available at \url{https://github.com/Ahus-AIM/lett-next}.
\end{abstract}
\keywords{Computed Tomography \and Interactive Tumor Segmentation}

%% file: sections/introduction.tex
\section{Introduction}

RECIST v1.1~\cite{eisenhauerNewResponseEvaluation2009} is widely used to standardize tumor response assessment in clinical practice and clinical trials. Its diameter-based measurements are practical and easy to acquire, but a small number of 2D measurements cannot fully represent the 3D shape and spatial extent of a lesion. Volumetric lesion segmentation can describe tumor burden more completely, and previous studies suggest that volumetric analysis may provide clinically useful information beyond RECIST in settings such as lung cancer response assessment and breast cancer treatment response prediction~\cite{hayesComparisonCTVolumetric2016,machadoPromptableCTFoundation2025}. However, full 3D lesion annotation is difficult to scale in time-constrained clinical workflows.

The CVPR 2026: Foundation Models for Pan-cancer Segmentation in CT Images competition addresses this gap by asking models to predict a 3D lesion mask from a CT volume and a RECIST marker. This setting requires accurate local segmentation from sparse spatial input, while lesions vary in size, shape, appearance, and anatomical location. The challenge also emphasizes efficient inference under strict computational constraints, making runtime and memory use important design considerations.

\subsection{Related work}

Medical image segmentation has long been dominated by encoder--decoder architectures, most notably U-Net~\cite{ronnebergerUNetConvolutionalNetworks2015}. 3D variants such as SegResNet~\cite{myronenko3DMRIBrain2018} extend this family to volumetric segmentation, while nnU-Net~\cite{isenseeNnUNetSelfadaptingFramework2018} showed the importance of adapting preprocessing, training, inference, and postprocessing to each dataset. U-Net-style models therefore remain strong baselines for medical segmentation, especially when combined with task-specific configuration~\cite{maFastLowResourceAccurate2026}.

Recent ConvNeXt-style models further show that convolutional backbones remain competitive for 3D medical image segmentation. MedNeXt~\cite{royMedNeXtTransformerDrivenScaling2023} adapts ConvNeXt~\cite{liuConvNet2020s2022} principles to volumetric segmentation through a fully convolutional encoder--decoder with residual upsampling and downsampling blocks. MedNeXt-v2~\cite{royMedNeXtv2Scaling3D2025} extends this line of work by scaling 3D ConvNeXt backbones for supervised representation learning and adding volumetric global response normalization. Its evaluation against several strong public segmentation models~\cite{wasserthalTotalSegmentatorRobustSegmentation2023,hantzeMRSegmentatorMultiModalitySegmentation2025,heVISTA3DUnifiedSegmentation2024,duSegVolUniversalInteractive2025,huangSTUNetScalableTransferable2023} makes it a useful reference point for modern 3D medical segmentation backbones.

Promptable segmentation conditions prediction models on user- or task-provided spatial information. General segmentation models established this paradigm for natural images, videos, and concept-level prompts~\cite{kirillovSegmentAnything2023,raviSAM2Segment2024,carionSAM3Segment2026}, while medical variants adapt promptable segmentation to medical images and volumetric data~\cite{maSegmentAnythingMedical2024,wangSAMMed3DGeneralpurposeSegmentation2024,duSegVolUniversalInteractive2025}. In 3D medical segmentation, nnInteractive is especially relevant because it represents spatial prompts as additional input channels and introduces AutoZoom for adaptive field-of-view refinement~\cite{isenseeNnInteractiveRedefining3D2025}.

ENSAM introduced a promptable 3D medical image segmentation framework with a SegResNet-based image encoder, prompt encoder, and mask decoder connected through latent cross-attention~\cite{stenhedeENSAMEfficientFoundation2025}. Lite ENSAM adapted this framework to RECIST-conditioned CT lesion segmentation by using RECIST markers as sparse prompts and by reducing the computational footprint for CPU-based inference~\cite{bjornstadLiteENSAMLightweight2025}. Together, these models provide the main starting point for the present work: prompt-conditioned volumetric segmentation, RECIST-guided lesion localization, and lightweight inference. 

Building on ENSAM and Lite ENSAM, we present LETT-NeXt, a lightweight RECIST-guided 3D lesion segmentation model. LETT-NeXt replaces latent prompt decoding with direct RECIST prompt-channel conditioning and uses a compact MedNeXt-v2 encoder–decoder for efficient local segmentation. During training, an auxiliary anatomy–tumor head provides additional supervision. At inference, RECIST-guided regional crops and AutoZoom maintain a local field of view, while still allowing adaptive expansion when needed.


%% file: sections/methods.tex
\section{Method}
The method section first describes RECIST-guided regional cropping and prompt encoding, then introduces the MedNeXt-v2 segmentation model, and finally describes prompt-aware postprocessing and AutoZoom inference.

\subsection{Preprocessing}
\label{sec:input_preprocessing}

All CT volumes were resampled to a spacing of \(2.4 \times 1.0 \times 1.0\) mm in \(z,y,x\) order. For each RECIST annotation, we extracted a RECIST-centered regional crop of size \(72 \times 160 \times 160\) voxels. Predictions produced in crop space were mapped back to the original image grid after inference.

CT intensities were clipped to \([-999,255]\) HU and normalized by subtracting the training-set mean \(\mu=69.86\) and dividing by the training-set standard deviation \(\sigma=200.88\). The normalized crop is denoted by \(\tilde{\mathbf{I}}_{\mathrm{CT}}\).

LETT-NeXt conditions the segmentation model on RECIST annotations by concatenating two prompt channels with the normalized CT crop. The first channel, \(\mathbf{P}_{\mathrm{line}}\), is a binary RECIST-line mask. The two RECIST endpoints are connected on the axial RECIST slice and thickened with disks of radius \(r=2\) voxels. Voxels covered by the thickened line are assigned value 1, and all other voxels are assigned value 0.

The second channel, \(\mathbf{P}_{\mathrm{endpoints}}\), is an endpoint heatmap. For voxel coordinate \(\mathbf{v}=(z,y,x)\) and RECIST endpoints \(\mathbf{p}_1,\mathbf{p}_2\) in crop coordinates, the heatmap is defined as
\[
\mathbf{P}_{\mathrm{endpoints}}(\mathbf{v})
=
\max_{i \in \{1,2\}}
\exp
\left(
-\frac{
\left\| \mathbf{v} - \mathbf{p}_i \right\|_2^2
}{
2\sigma_{\mathrm{end}}^2
}
\right),
\]
where \(\sigma_{\mathrm{end}}=2.0\) voxels controls the spatial spread of the endpoint prompt.

The final network input is the three-channel tensor
\[
\mathbf{X}
=
\left[
\tilde{\mathbf{I}}_{\mathrm{CT}},
\mathbf{P}_{\mathrm{line}},
\mathbf{P}_{\mathrm{endpoints}}
\right].
\]

\subsection{Proposed Method}

\begin{figure}[t]
\centering
\includegraphics[width=\textwidth]{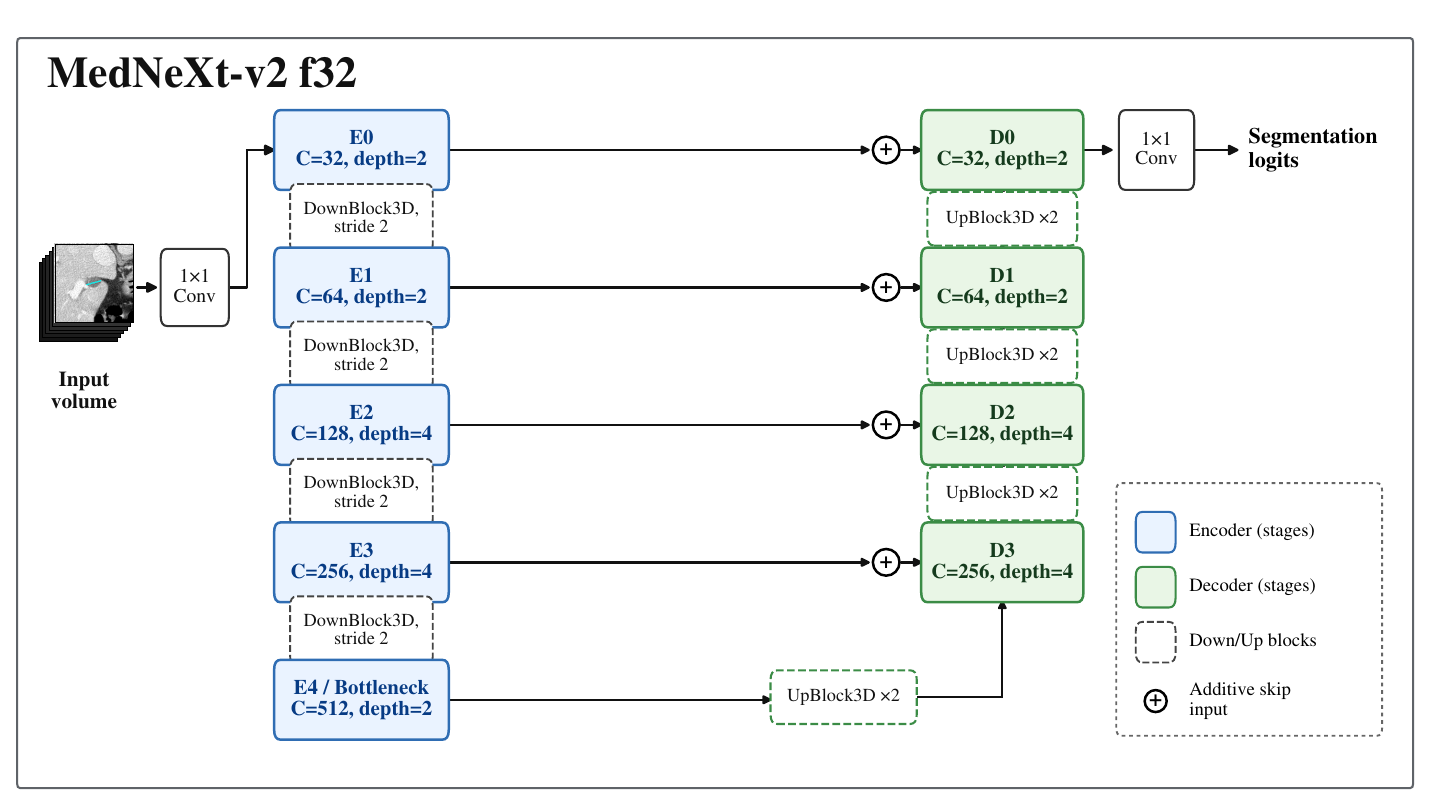}
\caption{
Overview of the MedNeXt-v2 f32 architecture used in LETT-NeXt.
The network has five encoder scales, four decoder scales, skip connections between corresponding resolution levels, and encoder channel widths of 32, 64, 128, 256, and 512.
}
\label{fig:mednext_architecture}
\end{figure}

Figure~\ref{fig:mednext_architecture} illustrates the LETT-NeXt segmentation network.
LETT-NeXt uses the three-channel input \(\mathbf{X}\) from Section~\ref{sec:input_preprocessing} and predicts a binary lesion mask with a compact MedNeXt-v2 f32 encoder--decoder~\cite{royMedNeXtv2Scaling3D2025}.
The network follows a 3D U-Net-like topology with skip connections between corresponding resolution levels and contains approximately 6.92 million trainable parameters.
Additional block-level details are provided in Appendix~\ref{app:mednext_blocks}.

\subsubsection{Auxiliary anatomy supervision}
\label{sec:auxiliary_anatomy_supervision}

During training, LETT-NeXt used a training-only auxiliary anatomy--tumor head in addition to the main RECIST-guided lesion segmentation head. This was motivated by anatomy-aware supervision in PanTS~\cite{liPanTSPancreaticTumor2025a}, and was used to encourage anatomy-aware feature learning around the prompted lesion.

The auxiliary head was implemented as a \(1 \times 1 \times 1\) convolution attached to the final decoder feature map. It predicted a local anatomy--tumor map for PanTS-supervised training crops, while the main RECIST-guided head predicted a binary lesion mask.

The auxiliary loss was computed only over voxels with valid PanTS-derived auxiliary labels. It combined multi-class cross-entropy and multi-class Dice loss over the auxiliary anatomy--tumor label space:
\[
\mathcal{L}_{\mathrm{aux}}
=
0.5\,\mathcal{L}_{\mathrm{CE}}
\left(
\hat{\mathbf{Y}}_{\mathrm{aux}},
\mathbf{Y}_{\mathrm{aux}}
\right)
+
0.5\,\mathcal{L}_{\mathrm{Dice}}^{\mathrm{multi}}
\left(
\hat{\mathbf{Y}}_{\mathrm{aux}},
\mathbf{Y}_{\mathrm{aux}}
\right).
\]
For crops without valid auxiliary labels, the auxiliary term was set to zero. At inference, only the main lesion head was used, and the auxiliary logits were ignored.

\subsection{Postprocessing}
\label{sec:inference_autozoom}
LETT-NeXt performed inference independently for each RECIST marker in a CT scan. 
Each prompt corresponds to one target lesion, so the pipeline produced one lesion prediction per RECIST marker. 
For each prompt, the corresponding input tensor \(\mathbf{X}\) was passed through the MedNeXt-v2 f32 model.
The sigmoid output was thresholded at 0.35 to obtain candidate lesion voxels. This value was selected by a validation-set threshold sweep.

To ensure that each prediction corresponded to a single prompted lesion, connected components were computed on the thresholded crop-space mask. 
The component most consistent with the RECIST marker was retained. 
Selection first used the component with the largest overlap with the RECIST marker region. 
If no component overlapped the prompt region, the component closest to the prompt center was selected instead. 
The selected crop-space mask was mapped back to the original image grid and assigned the corresponding RECIST label ID in the final output mask.

\subsubsection{AutoZoom}

AutoZoom is designed for cases where the initial RECIST-centered crop may not fully contain the lesion. Small crops preserve more image detail, but lesions that extend beyond the crop boundary can be truncated. Following the prompt-guided local inference strategy of nnInteractive~\cite{isenseeNnInteractiveRedefining3D2025}, LETT-NeXt performs a second inference pass only when the first-pass prediction touches the crop boundary.

For the second pass, a larger physical crop is extracted around the same RECIST marker, resized to the fixed network input size, and passed through the same model. The expanded prediction replaces the initial prediction, after which the same thresholding, component selection, and mapping procedure is applied.

%% file: sections/experiments.tex
\section{Experiments}

\subsection{Dataset and evaluation measures}
\label{sec:dataset_evaluation}

The development data were provided by the CVPR 2026: Foundation Models for Pan-cancer Segmentation in CT Images competition and hosted on Hugging Face~\cite{FLAREMedFMFLARETask1PancancerRECISTto3DDatasets}. The challenge-provided training and validation partitions are referred to as the training and validation sets.

The training data contained 25,112 records in total: 24,311 primary challenge records and 801 PanTS records used only for auxiliary supervision. The primary records came from 9,968 unique cases. Training was performed at the lesion-prompt level: each competition record paired one CT volume with one target lesion mask, voxel spacing metadata, and the corresponding RECIST marker. Therefore, cases with multiple target lesions could contribute multiple training records. PanTS records were used only for auxiliary anatomy--tumor supervision.

The validation set contained 49 CT cases with 84 annotated target lesions and RECIST markers. At inference, prompt-level predictions were merged into one labeled output mask per CT case. Unless otherwise stated, validation metrics were computed per CT case and reported as mean \(\pm\) SD over the 49 cases. Hidden test results were obtained from the challenge evaluation server, which reported aggregate metrics only.

Development results were measured using the challenge score, defined as the mean of Dice Similarity Coefficient (DSC) and Normalized Surface Dice (NSD):
\begin{equation}
\label{eq:score}
    \mathrm{Score}
    =
    0.5 \cdot \mathrm{DSC}
    +
    0.5 \cdot \mathrm{NSD}.
\end{equation}
Unless otherwise stated, DSC, NSD, and challenge score are reported on a \(0\)--\(100\) scale, obtained by multiplying the corresponding \([0,1]\) metric values by 100. Absolute differences are reported in percentage points.

\subsection{Implementation Details}
\label{sec:implementation_details}

LETT-NeXt was trained using a mixed RECIST-centered crop recipe. Standard crops were centered on the RECIST marker and endpoints, matching the prompt geometry used at inference. Enlarged-field crops used the same center but sampled a larger physical field of view before resizing to the fixed input size. Bounding-box-fitted crops were used only when a standard RECIST-centered crop risked cutting off part of the lesion; in these cases, the crop was centered on the lesion bounding box and adjusted to keep the full lesion inside the patch.

The nominal sampling recipe was 60\% standard RECIST marker crops, 30\% enlarged-field RECIST crops, and up to 10\% bounding-box-fitted crops. If bounding-box-fitted sampling was not applicable, sampling fell back to the standard RECIST marker crop. All crop types enforced inclusion of the RECIST endpoints with a \SI{15}{\milli\metre} margin.

Table~\ref{tab:training_protocol} summarizes the training protocol. Hardware and runtime details for the final training run are reported in Appendix~\ref{app:implementation_details}, Table~\ref{tab:hardware_appendix}.

\begin{table}[!htbp]
\centering
\small
\caption{Training protocol for LETT-NeXt.}
\label{tab:training_protocol}
\begin{tabular}{p{0.38\linewidth} p{0.48\linewidth}}
\toprule
\textbf{Parameter} & \textbf{Value} \\
\midrule
Backbone & MedNeXt-v2 f32 \\
Input crop size & \(72 \times 160 \times 160\) voxels, \(z,y,x\) \\
Target spacing & \(2.4 \times 1.0 \times 1.0\) mm, \(z,y,x\) \\
Training data & 9,968 cases; 25,112 records \\
Validation set & 49 CT cases; 84 RECIST markers \\
Epochs & 25 \\
Optimizer & AdamW \\
Initial learning rate & \(10^{-3}\) \\
Batch size & 6 per GPU, 18 global \\
Image augmentations & Intensity shift 0.03, intensity scaling 0.05, Gaussian noise \(\sigma=0.01\) \\
\bottomrule
\end{tabular}
\end{table}

The main lesion loss was defined as
\begin{equation}
    \mathcal{L}_{\mathrm{main}}
    =
    \mathcal{L}_{\mathrm{Dice}}
    +
    2\mathcal{L}_{\mathrm{BCE}}.
\end{equation}

The total training loss used the auxiliary supervision loss described in Section~\ref{sec:auxiliary_anatomy_supervision}:
\begin{equation}
    \mathcal{L}_{\mathrm{total}}
    =
    \mathcal{L}_{\mathrm{main}}
    +
    0.25\,\mathcal{L}_{\mathrm{aux}}.
\end{equation}

%% file: sections/results.tex
\section{Results}
\label{sec:results}
LETT-NeXt was evaluated in terms of segmentation accuracy, qualitative behavior, and computational efficiency. 
Quantitative performance was measured on the public validation set and hidden test set using DSC, NSD, and the challenge score. Metrics are reported on a 0--100 scale, and absolute differences are reported in percentage points.

\subsection{Quantitative validation results}

LETT-NeXt completed all 49 public validation cases without failures or timeouts. 
Table~\ref{tab:final-results} summarizes public validation and hidden test performance. 
Compared with the Lite ENSAM competition baseline, LETT-NeXt achieved higher hidden test DSC, NSD, and challenge score. 
On public validation, LETT-NeXt obtained higher DSC but lower NSD, resulting in a lower overall validation score.

\begin{table}[htbp]
\caption{Quantitative evaluation results on the public validation and hidden test sets. The LETT-NeXt validation result uses the inference pipeline with AutoZoom enabled.}
\label{tab:final-results}
\centering
\begin{tabular}{lcccccc}
\toprule
\multirow{2}{*}{Model} & \multicolumn{3}{c}{Public Validation} & \multicolumn{3}{c}{Hidden Test} \\
\cmidrule(lr){2-4} \cmidrule(lr){5-7}
                         & DSC(\%) & NSD(\%) & Score(\%) & DSC(\%) & NSD(\%) & Score(\%) \\
\midrule
Lite ENSAM               & $76.10 \pm 16.30$ & $78.90 \pm 19.10$ & $77.50$            & $60.70$ & $63.60$ & $62.15$ \\
LETT-NeXt                & $79.43 \pm 10.09$ & $72.32 \pm 16.22$ & $75.87 \pm 12.17$ & $73.90$ & $67.30$ & $70.60$ \\
\bottomrule
\end{tabular}
\end{table}

The larger standard deviation for NSD suggests that boundary accuracy varied more than volumetric overlap. 
This is expected in RECIST-guided segmentation, because the RECIST marker provides lesion position and approximate scale, but not the full 3D lesion boundary.

\subsection{Qualitative results}

\begin{figure}[htbp]
    \centering
    \includegraphics[width=\linewidth]{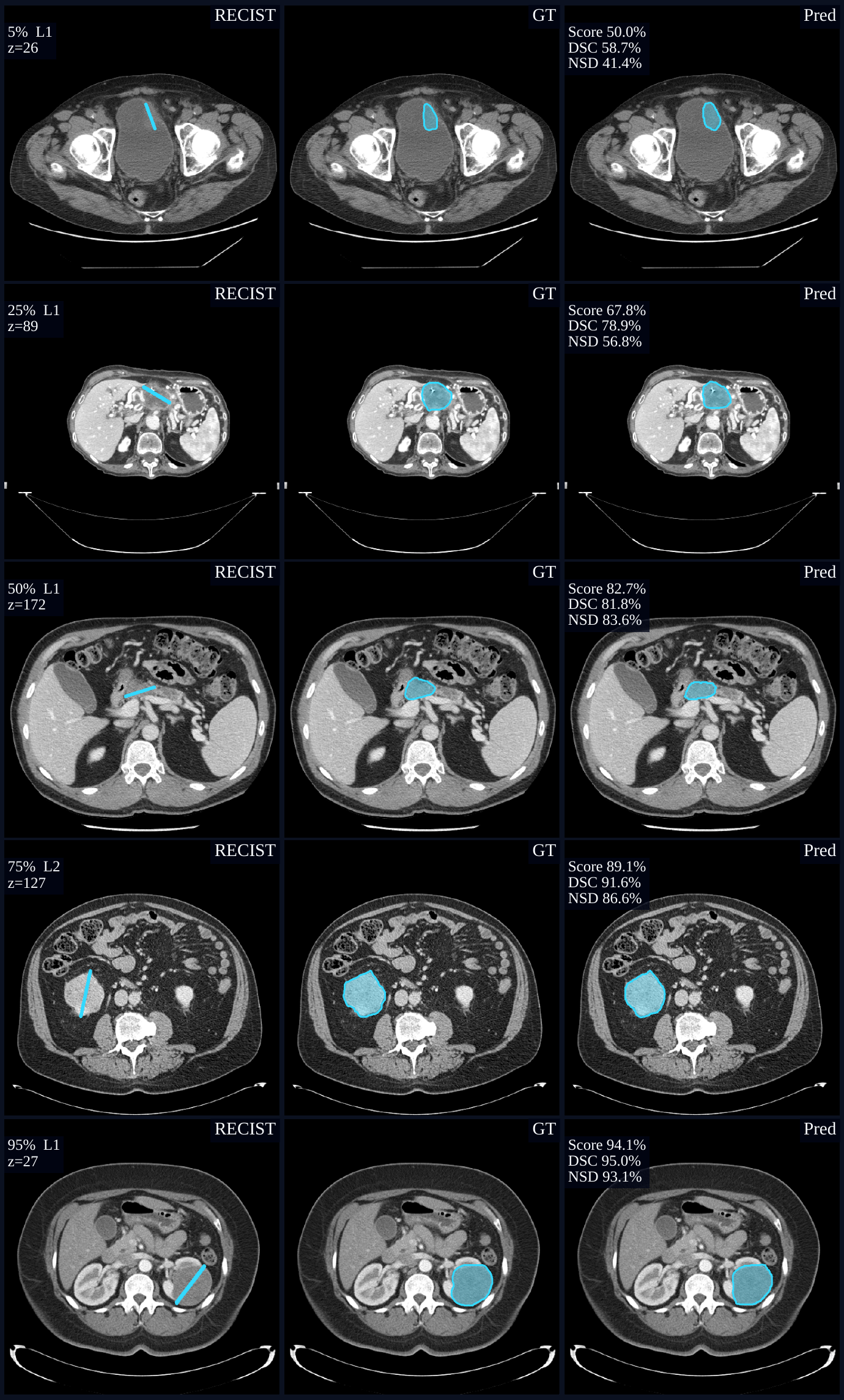}
    \caption{
    Qualitative validation examples across score percentiles. 
    Each row shows the RECIST input marker, ground-truth mask, and LETT-NeXt prediction on the prompt slice. 
    Higher-score examples generally show stronger agreement with the ground truth. 
    }
    \label{fig:percentile_examples}
\end{figure}

\Cref{fig:percentile_examples} shows representative validation examples selected by score percentile. Qualitatively, LETT-NeXt usually localized the prompted lesion, but the main visible failure mode was under-segmentation. In difficult cases, the model predicted masks that were too small, especially for small lesions, low-contrast lesions, and lesions with unclear boundaries.

\subsection{Runtime and memory}

LETT-NeXt satisfied the CPU inference constraints on the 49-case public validation set. 
Table~\ref{tab:efficiency_results} summarizes validation set runtime and memory use. 
The hidden test server reported a mean runtime of \(20.9\,\mathrm{s}\) per case under hidden challenge-server conditions.

\begin{table}[htbp]
    \centering
    \caption{Runtime and memory use on the 49-case public validation set.}
    \label{tab:efficiency_results}
    \begin{tabular}{ll}
        \toprule
        \textbf{Quantity} & \textbf{Value} \\
        \midrule
        Mean runtime per case & \(6.88 \pm 3.00\,\mathrm{s}\) \\
        Maximum runtime & \(13.55\,\mathrm{s}\) \\
        Mean CPU memory-time & \(11.01 \pm 5.49\,\mathrm{GB\,s}\) \\
        Total CPU memory-time & \(539.36\,\mathrm{GB\,s}\) \\
        Maximum resident memory & \(3.57\,\mathrm{GB}\) \\
        \bottomrule
    \end{tabular}
\end{table}

%% file: sections/ablations.tex
\section{Ablation Studies}
\label{sec:ablation_studies}

Ablation studies evaluated three components of the RECIST-guided lesion segmentation pipeline: backbone architecture, auxiliary anatomy supervision, and adaptive field-of-view expansion with AutoZoom. These controlled ablations used shorter training runs and smaller crops than the final model, so they should be interpreted as component comparisons rather than direct final-model results. Additional setup details are provided in Appendix~\ref{app:controlled_ablation_details}.

\subsection{Architecture Ablation}
\label{sec:architecture_ablation}

Five segmentation backbones were compared using the controlled protocol described in Appendix~\ref{app:controlled_ablation_details}. Auxiliary anatomy supervision was disabled for this comparison.

Table~\ref{tab:architecture_ablation_main} shows the 49-case validation results. MedNeXt-v2 f32 achieved the best validation score, with \(71.40\), compared with \(65.08\) for the next-best backbone, SwinUNETR f24. In the paired comparison, this corresponded to an improvement of \(+6.32\) percentage points, with a 95\% bootstrap CI of \([+3.90,+8.80]\) percentage points. MedNeXt-v2 f32 was therefore selected as the final segmentation backbone.

\begin{table}[htbp]
\centering
\caption{
Matched architecture ablation on the 49-case validation set.
}
\label{tab:architecture_ablation_main}
\begin{tabular}{lcccc}
\toprule
\textbf{Backbone} & \textbf{Params.} & \textbf{Score (\%)} & \textbf{DSC (\%)} & \textbf{NSD (\%)} \\
\midrule
UNet base         & 1.19M  & 63.90 & 70.87 & 56.92 \\
SegResNet         & 4.70M  & 63.59 & 72.13 & 55.06 \\
SwinUNETR f24     & 15.70M & 65.08 & 72.32 & 57.84 \\
MedNeXt-v2 small  & 1.28M  & 65.04 & 72.16 & 57.92 \\
\textbf{MedNeXt-v2 f32}
                  & \textbf{6.92M}
                  & \textbf{71.40}
                  & \textbf{76.06}
                  & \textbf{66.74} \\
\bottomrule
\end{tabular}
\end{table}

\subsection{Auxiliary Anatomy Supervision}
\label{sec:auxiliary_head_ablation}

The auxiliary head was evaluated using a matched 15-epoch control experiment described in Appendix~\ref{app:auxiliary_head_control}. 
The comparison tested whether enabling the PanTS auxiliary anatomy--tumor head improved RECIST-guided lesion segmentation.

Table~\ref{tab:aux_head_ablation_main} reports the best-threshold validation results. 
Auxiliary supervision improved the mean validation score by \(+0.96\) percentage points, but the 95\% bootstrap confidence interval included zero \([-0.18,+2.05]\). 
The auxiliary head is therefore interpreted as a weak training-time regularizer rather than a definitive improvement or standalone anatomy segmentation component.

\begin{table}[htbp]
\centering
\caption{
Auxiliary-head ablation using a matched MedNeXt-v2 small prompt setup.
Metrics are reported as mean \(\pm\) SD over the 49-case validation set on a \(0\)--\(100\) scale.
}
\label{tab:aux_head_ablation_main}
\begin{tabular}{lcccc}
\toprule
\textbf{Variant} & \textbf{Threshold} & \textbf{Score (\%)} & \textbf{DSC (\%)} & \textbf{NSD (\%)} \\
\midrule
No auxiliary head
& 0.35
& \(66.13 \pm 13.09\)
& \(72.63 \pm 11.18\)
& \(59.63 \pm 18.45\) \\
PanTS auxiliary head
& 0.35
& \(\mathbf{67.09 \pm 13.29}\)
& \(\mathbf{73.39 \pm 11.08}\)
& \(\mathbf{60.80 \pm 18.68}\) \\
\midrule
Paired improvement
& --
& \(+0.96 \pm 4.01\)
& \(+0.76 \pm 3.15\)
& \(+1.17 \pm 5.21\) \\
\bottomrule
\end{tabular}
\end{table}

Additional threshold-sweep results are provided in Appendix~\ref{app:auxiliary_head_control}.

\subsection{AutoZoom}
\label{sec:autozoom_ablation}

Table~\ref{tab:autozoom_ablation} compares fixed RECIST-centered regional crop inference with adaptive AutoZoom. 
The table reports the mean challenge score on a \(0\)--\(100\) scale, where \(\Delta\) denotes AutoZoom minus fixed-crop inference in percentage points. 
AutoZoom triggered in only a small fraction of cases and had near-neutral aggregate performance. 
On the validation set, the difference was small, with a 95\% bootstrap CI spanning zero \([-0.32,+0.02]\) percentage points. 
AutoZoom is therefore best interpreted as a selective fallback for cases where the initial RECIST-centered crop may truncate the target lesion.

\begin{table}[htbp]
\centering
\caption{
AutoZoom ablation. 
Scores denote the mean challenge score on a \(0\)--\(100\) scale. 
\(\Delta\) is reported as AutoZoom minus fixed-crop inference in percentage points.
``Largest'' refers to the largest PanTS targets by lesion size.
}
\label{tab:autozoom_ablation}
\setlength{\tabcolsep}{5pt}
\renewcommand{\arraystretch}{1.05}
{\small
\begin{tabular}{lrrrrr}
\toprule
\multirow{2}{*}{\textbf{Subset}} 
& \multirow{2}{*}{\textbf{\(N\)}} 
& \multirow{2}{*}{\textbf{Triggers}} 
& \multicolumn{2}{c}{\textbf{Challenge score (\%)}} 
& \multirow{2}{*}{\textbf{\(\Delta\) (pp)}} \\
\cmidrule(lr){4-5}
& & & \textbf{Fixed crop} & \textbf{AutoZoom} & \\
\midrule
Largest 10     & 10  & 1 \; (10.0\%) & 62.46 & 63.51 & +1.05 \\
Largest 40     & 40  & 5 \; (12.5\%) & 57.03 & 58.05 & +1.02 \\
Largest 120    & 120 & 7 \; (5.8\%)  & 63.91 & 64.34 & +0.43 \\
All PanTS      & 801 & 23 \; (2.9\%) & 64.10 & 64.19 & +0.09 \\
Validation set & 49  & 3 \; (6.1\%)  & 75.99 & 75.87 & -0.12 \\
\bottomrule
\end{tabular}
}
\end{table}

Adaptive expansion was retained to improve robustness in context-limited cases rather than to increase average validation performance. 
The largest-lesion subsets showed small positive changes, while the full validation set showed a slight decrease. 
These results suggest that AutoZoom can recover missed lesion extent when additional field of view is needed, but may also reduce segmentation quality when the expanded crop lowers the effective spatial resolution.

\begin{figure}[htbp]
    \centering
    \includegraphics[width=\linewidth]{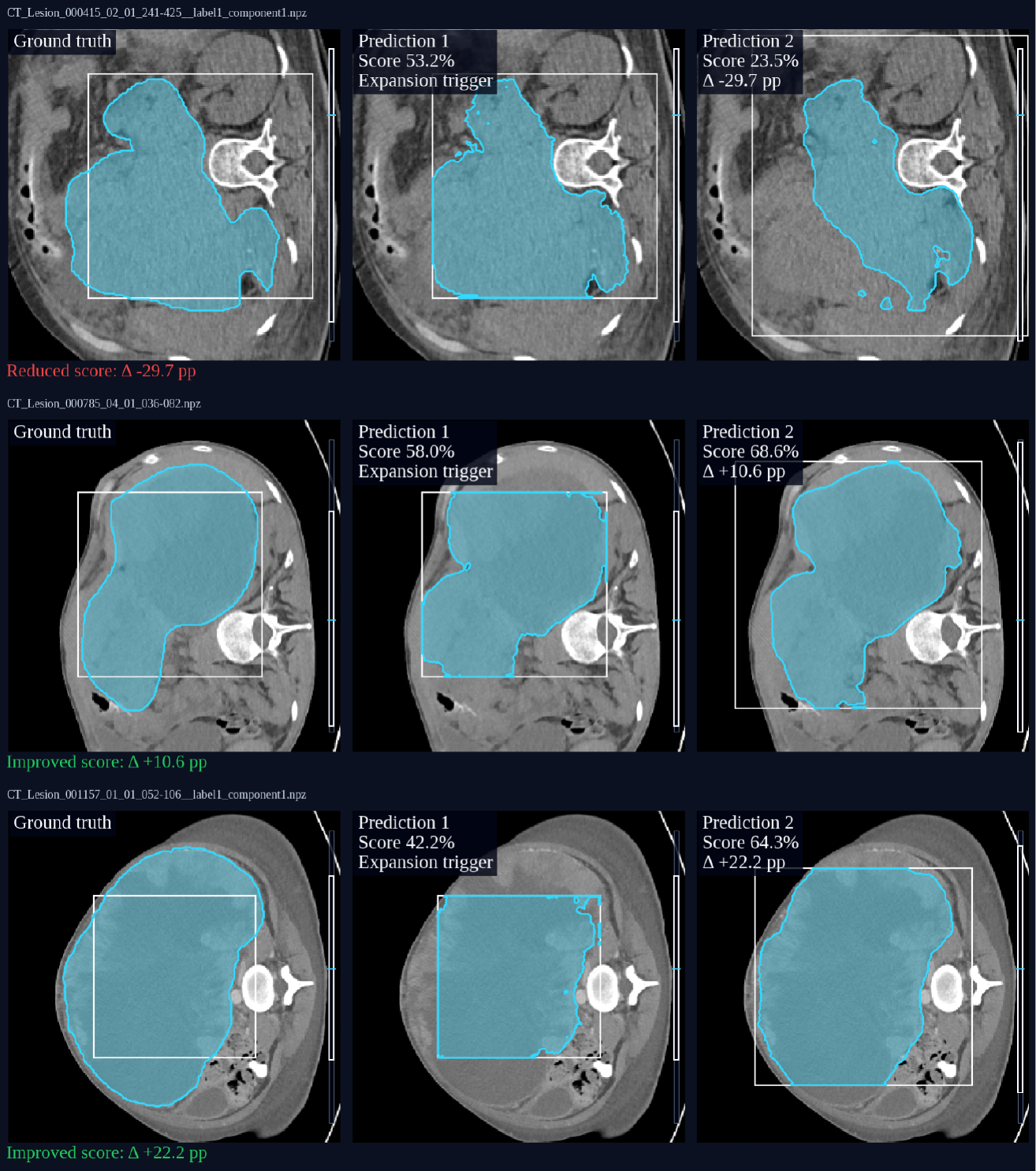}
    \caption{
    Qualitative examples of RECIST-guided AutoZoom on the validation set.
    Each row shows the ground truth, the first prediction using the initial crop, and the second prediction after adaptive crop expansion.
    }
    \label{fig:autozoom_examples}
\end{figure}

Figure~\ref{fig:autozoom_examples} shows qualitative validation examples. 
Adaptive expansion recovered missed lesion extent in some cases, but reduced segmentation quality in others.

%% file: sections/discussion.tex
\section{Discussion}

LETT-NeXt demonstrates that sparse RECIST markers can support efficient 3D lesion segmentation under CPU inference constraints. 
The model generally localized the marked lesions, while the remaining errors mainly involved lesion extent and boundary placement. 
This pattern is consistent with the lower NSD than DSC on the public validation set and with the under-segmentation examples in \Cref{fig:percentile_examples}. 
In practice, this suggests that the RECIST marker provides strong localization information, but does not fully resolve the 3D lesion boundary.

The controlled ablations suggest that backbone choice was the main driver of performance. 
The MedNeXt-v2 f32 backbone provided the clearest performance gain, whereas auxiliary supervision acted more like weak training-time regularization than a definitive improvement. 
AutoZoom was neutral on average, but remained useful as a selective fallback for cases where the initial regional crop may not fully contain the lesion.

The main limitations are the small 49-case validation set, threshold selection on the same local validation split used for model development, limited hidden-test detail, and the absence of external validation. 
The ablations also tested components mostly in isolation, so interactions between backbone choice, auxiliary supervision, crop strategy, and postprocessing remain uncertain. 
Future work should evaluate LETT-NeXt on external datasets and improve boundary refinement. 
A promising direction is to pre-train a larger anatomy-aware model on dense multi-class anatomy--tumor labels, for example, from PanTS, before fine-tuning it for RECIST-marker tumor segmentation or distilling it into a compact student model. 
This strategy could improve anatomical context and boundary modeling while preserving efficient CPU inference.

Overall, LETT-NeXt shows that RECIST markers can support accurate and efficient local 3D lesion segmentation, but further work is needed to improve boundary accuracy and robustness across datasets.

\section*{Acknowledgements}
The authors express their appreciation to Novartis Norge AS and Akershus University Hospital for funding this work. The submitted method for the \emph{Foundation Models for Pan-cancer Segmentation in CT Images} competition is fully automatic and does not require manual intervention. Only data permitted by the organizers was used. We thank the challenge organizers, data providers, and CodaBench~\cite{xuCodabenchFlexibleEasytoUse2022} for hosting the competition.

%% file: sections/appendix.tex
\appendix

\section{Implementation Details}
\label{app:implementation_details}

Training used distributed data parallelism on three NVIDIA GeForce RTX 5090 GPUs.
Table~\ref{tab:hardware_appendix} summarizes the hardware and runtime used for the final training run.

\begin{table}[htbp]
\centering
\small
\caption{Hardware and runtime used for the final training run.}
\label{tab:hardware_appendix}
\begin{tabular}{p{0.38\linewidth} p{0.48\linewidth}}
\toprule
\textbf{Item} & \textbf{Value} \\
\midrule
Operating system & Linux 6.1.0-44-amd64 \\
CPU & 24 physical cores / 48 threads \\
System memory & 202 GB RAM \\
GPUs & 3 $\times$ NVIDIA GeForce RTX 5090 \\
GPU memory & 34 GB per GPU \\
Training parallelism & Distributed data parallel \\
Data-loader workers & 8 per process \\
Training runtime & 25.2 h for the complete training \\
\bottomrule
\end{tabular}
\end{table}

\section{MedNeXt-v2 Building Blocks}
\label{app:mednext_blocks}

\begin{figure}[htbp]
\centering
\includegraphics[
    width=\textwidth,
    trim={0 0 0 0.5cm},
    clip
]{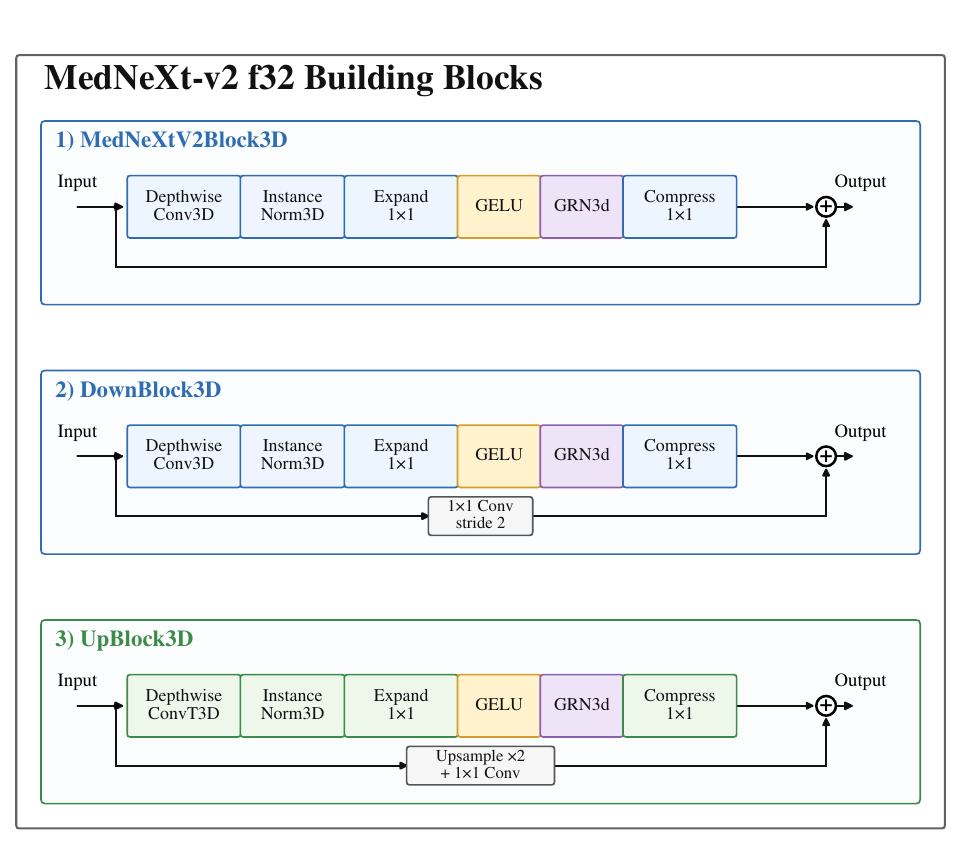}
\caption{
Core MedNeXt-v2 building blocks used in LETT-NeXt.
The main block uses depthwise convolution, normalization, pointwise expansion and compression, GELU activation, global response normalization, and a residual connection.
The downsampling and upsampling blocks use residual projections to change spatial resolution.
}
\label{fig:mednext_blocks}
\end{figure}

Figure~\ref{fig:mednext_blocks} summarizes the core MedNeXt-v2 blocks used in LETT-NeXt.

\section{Controlled Ablation Details}
\label{app:controlled_ablation_details}

This appendix provides the shared protocol for the controlled ablations reported in Section~\ref{sec:ablation_studies}. The architecture ablation and auxiliary-head control used the same training data, crop geometry, prompt encoding, training schedule, validation set, and postprocessing unless otherwise stated. Paired improvements were computed from per-case validation-score differences on the 49-case validation set. Uncertainty was estimated using 95\% percentile bootstrap confidence intervals over validation cases.

\subsection{Shared Training and Validation Setup}
\label{app:shared_ablation_setup}

All controlled ablation models were trained for 15 epochs with 625 steps per epoch and a global batch size of 8. This corresponds to 5,000 sampled crop presentations per epoch and 75,000 crop presentations in total. The crop size was \(64 \times 128 \times 128\) voxels and the target spacing was \(2.4 \times 1.0 \times 1.0\) mm, both given in \(z,y,x\) order.

The training data contained 25,112 records: 24,311 primary training records and 801 PanTS records. Validation was performed on the 49-case validation set. All models used the same three-channel input formulation defined in Section~\ref{sec:input_preprocessing}. The main output was a single binary lesion logit. The main lesion loss was Dice loss plus BCE loss with weights 1:2. Validation used prompt-component postprocessing, AutoZoom, threshold \(0.5\) unless otherwise stated, and cached dataset-statistics normalization.

\subsection{Architecture Ablation}
\label{app:architecture_ablation_details}

The architecture ablation compared five segmentation backbones under the shared setup described in Section~\ref{app:shared_ablation_setup}. The only intended experimental variable was the backbone architecture. Auxiliary anatomy supervision was disabled for all models in this ablation.

Table~\ref{tab:architecture_ablation_backbones} summarizes the backbones included in the comparison. All models used the same three-channel input formulation and binary lesion output, keeping the prompt representation fixed across backbones. The RECIST marker and endpoint maps were concatenated with the CT crop, so each backbone received the prompt as voxel-aligned image information.

\begin{table}[t]
\centering
\caption{
Overview of the backbones included in the architecture ablation.
}
\label{tab:architecture_ablation_backbones}
\footnotesize
\setlength{\tabcolsep}{4pt}
\begin{tabularx}{\textwidth}{@{}l@{\hspace{1.2em}}Xr@{}}
\toprule
\textbf{Model} & \textbf{Family} & \textbf{Params.} \\
\midrule
UNet base 
& 3D UNet 
& 1.19\,M \\

SegResNet 
& Residual CNN 
& 4.70\,M \\

SwinUNETR f24 
& Transformer--CNN hybrid 
& 15.70\,M \\

MedNeXt-v2 small 
& ConvNeXt-style 3D CNN 
& 1.28\,M \\

MedNeXt-v2 f32 
& ConvNeXt-style 3D CNN 
& 6.92\,M \\
\bottomrule
\end{tabularx}
\end{table}

\subsection{Auxiliary-Head Control}
\label{app:auxiliary_head_control}

The auxiliary-head control compared two MedNeXt-v2 small models under the shared setup described in Section~\ref{app:shared_ablation_setup}. The only intended experimental variable was whether the PanTS auxiliary anatomy--tumor head was enabled during training.

Table~\ref{tab:aux_head_threshold_sweep} reports the threshold sweep for the auxiliary-head control experiment. The auxiliary-head variant achieved a higher validation score at every swept threshold.

\begin{table}[t]
\centering
\caption{
Threshold sweep for the auxiliary-head control experiment.
Scores are reported on a \(0\)--\(100\) scale, and differences are reported in percentage points.
}
\label{tab:aux_head_threshold_sweep}
\begin{tabular}{cccc}
\toprule
\textbf{Threshold} & \textbf{No aux score (\%)} & \textbf{Aux score (\%)} & \textbf{\(\Delta\) (pp)} \\
\midrule
0.35 & 66.13 & \textbf{67.09} & +0.96 \\
0.40 & 65.65 & 67.02 & +1.37 \\
0.45 & 65.41 & 66.78 & +1.37 \\
0.50 & 65.02 & 66.36 & +1.34 \\
0.55 & 64.84 & 65.79 & +0.95 \\
0.60 & 64.49 & 65.45 & +0.95 \\
0.65 & 63.96 & 64.77 & +0.81 \\
0.70 & 63.34 & 64.02 & +0.69 \\
0.75 & 62.58 & 63.32 & +0.74 \\
\bottomrule
\end{tabular}
\end{table}

The effect of auxiliary supervision was likely limited by its small share of the training data: PanTS contributed 801 of 25,112 training records, corresponding to 3.2\%. The auxiliary branch also learned a tumor signal on the 50-case PanTS auxiliary evaluation set, reaching an auxiliary tumor DSC of \(62.38\%\). These results should be interpreted as a matched single-run control rather than a multi-seed estimate.